\PassOptionsToPackage{hyperfootnotes=false}{hyperref}
\PassOptionsToPackage{table}{xcolor}

\documentclass[times, review, 10pt]{elsarticle}

\biboptions{numbers,sort&compress}

\usepackage{amsmath}
\usepackage{amssymb}
\usepackage{amsthm}
\usepackage{algorithm}
\usepackage{algpseudocode}
\usepackage{enumitem}
\usepackage{bm}
\usepackage{pifont}
\usepackage{xcolor}
\usepackage{booktabs}
\usepackage{multirow}
\usepackage{array}
\usepackage{tikz}
\usepackage{tabularx}
\newcolumntype{Y}{>{\centering\arraybackslash}X}

\usepackage{setspace}
\usepackage[pdfencoding=auto]{hyperref}

\newcommand{\primtag}[1]{#1}
\newcommand{\casfloatfont}{\normalfont}

\usepackage[font={small,stretch=1.0},labelfont=bf,skip=6pt]{caption}

\usepackage{etoolbox}
\AtBeginEnvironment{table}{\setstretch{1.0}\small}
\AtBeginEnvironment{table*}{\setstretch{1.0}\small}
\AtBeginEnvironment{figure}{\setstretch{1.0}}
\AtBeginEnvironment{figure*}{\setstretch{1.0}}


\theoremstyle{plain}
\newtheorem{theorem}{Theorem}
\newtheorem{proposition}{Proposition}
\newtheorem{lemma}{Lemma}
\theoremstyle{definition}
\newtheorem{definition}{Definition}

\newcommand{\cmark}{\ding{51}}                 % supported
\newcommand{\xmark}{\ding{55}}                 % undefined / fails to run
\newcommand{\hmark}{\tikz[baseline=-0.55ex]{%  % degraded / unstable
  \draw[line width=.4pt] (0,0) circle (.45ex);
  \fill (0,-.45ex) arc (-90:90:.45ex) -- cycle;}}
\newcommand{\nd}{\textup{--}}                  % not applicable / not run
\begin{document}

\begin{frontmatter}

\title{HiLRP: Conservation-Valid Attribution for Hierarchical Vision
       Transformers via Attention Primitives}

\author[aff1]{Sathiyamohan Nishankar\corref{cor1}}
\ead{e17230@eng.pdn.ac.lk}

\author[aff3]{Pubudu Sanjeewani}

\author[aff4,aff5]{Asanka Perera}

\author[aff6]{Selvarajah Thuseethan}

\cortext[cor1]{Corresponding author.}

\affiliation[aff1]{organization={Faculty of Engineering,
                                University of Peradeniya},
                   country={Sri Lanka}}

\affiliation[aff3]{organization={School of Computing Technologies, RMIT University}, city={Melbourne}, country={Australia}}

\affiliation[aff4]{organization={School of Engineering \& Digital Technologies, University of Southern Queensland}, city={Brisbane}, country={Australia}}

\affiliation[aff5]{organization={School of Engineering \& Technologies, UNSW}, city={Canberra}, country={Australia}}

\affiliation[aff6]{organization={Faculty of Science and Technology, Charles Darwin University}, city={Darwin}, country={Australia}}

\begin{abstract}
Vision Transformer (ViT) architectures increasingly incoporate convolutional stems, windowed, linear, or multi-axis attention, patch merging, and spatial reduction in various configurations. This architectural diversity challenges existing attribution methods, whose assumptions often do not hold across ViT variants, Grad-CAM requires terminal spatial feature maps, attention rollout assumes global softmax attention, and layer-wise relevance propagation (LRP) requires module-specific rules. To the best of our knowledge, no existing attribution method provides a unified attribution framework in this architectural space. We show that this structural diversity can be captured by a simpler underlying foundation. The attention and resolution-reduction operators in current ViTs can be decomposed into four fundamental operation types: linear maps, bilinear mixing, normalization or gating, and reindexing. Each operation admits a relevance rule that satisfies conservation. Based on these rules, HiLRP supports new backbones by construction rather than by architecture-specific derivation, and its attribution maps decompose the prediction rather than relying on heuristic assumptions. We prove conservation and conditional equivariance, verifying both with respect to machine precision. Across 14 attribution methods and 10 architectures, we find that no prior method remains reliable across ViT families, while Faithfulness Correlation becomes uninformative for backbones robust to spatial masking. HiLRP alone preserves conservation across windowed, spatial-reduction, multi-axis, and linear-attention models, where naive extensions can produce zero or inflated relevance. It also overcomes localization failures in class activation mapping, achieving 0.97 Pointing compared to 0.55 for competing methods on EfficientViT. The same backward pass produces stage-localized maps, attributes label-free self-supervised objectives, and explains CLIP image-text similarity without requiring additional rules. The code is available at \url{https://github.com/Nishan-Charlie/Hi-LRP-Towards-One-Trustworthy-Explainable-AI.git}.
\end{abstract}

\begin{keyword}
Explainable artificial intelligence \sep
Vision transformers \sep
Layer-wise relevance propagation \sep
Faithfulness \sep
Localization \sep
Shapley agreement
\end{keyword}

\end{frontmatter}

\section{Introduction}
\label{sec:introduction}
Trust in artificial intelligence requires knowing not just what a model predicts, but why. As Vision Transformers (ViTs) increasingly dictate critical decisions in medical diagnosis, remote sensing, and industrial inspection, this "why" carries profound consequences. A clinician must know if a model has detected a genuine lesion or is simply focused on a scan artifact before acting on its finding. Post-hoc attribution methods attempt to provide this transparency by scoring each pixel's contribution to the output. However, for these scores to be trustworthy, the method must mathematically align with the underlying architecture and offer a precisely interpretable value. Currently, as ViT architectures rapidly evolve, neither condition is guaranteed.

This disparity arises because architecture design has rapidly evolved, whereas the attribution methodology has not. Most existing attribution frameworks were explicitly designed for the original ViT~\cite{dosovitskiy2021vit}, assuming a fixed token grid processed through global self-attention. However, contemporary backbones fundamentally restructure this paradigm. Swin~\cite{liu2021swin} restricts attention to shifted local windows, while PVT~\cite{wang2022pvtv2} contracts key and value grids prior to attention. Other variants introduce additional complexities: EfficientViT~\cite{cai2023efficientvit} substitutes the softmax operation with a linear cross-covariance product, MaxViT~\cite{tu2022maxvit} alternates between local and global attention, and MobileViT~\cite{mehta2022mobilevitv2} interleaves convolutional and transformer stages. Furthermore, spatial resolution is now routinely reduced across hierarchical stages via token merging or strided projections. Currently, innovations in training paradigms (DeiT~\cite{touvron2021deit}), pre-training objectives (BEiT~\cite{bao2022beit}), and modernized convolutional networks (ConvNeXt~\cite{liu2022convnext}) reflect an aggressive trajectory of structural optimization. Because this profound architectural diversification occurred within a mere three years, standard attribution frameworks have struggled to adapt.

\begin{figure}[!tbp]
  \centering
  \includegraphics[width=0.88\linewidth]{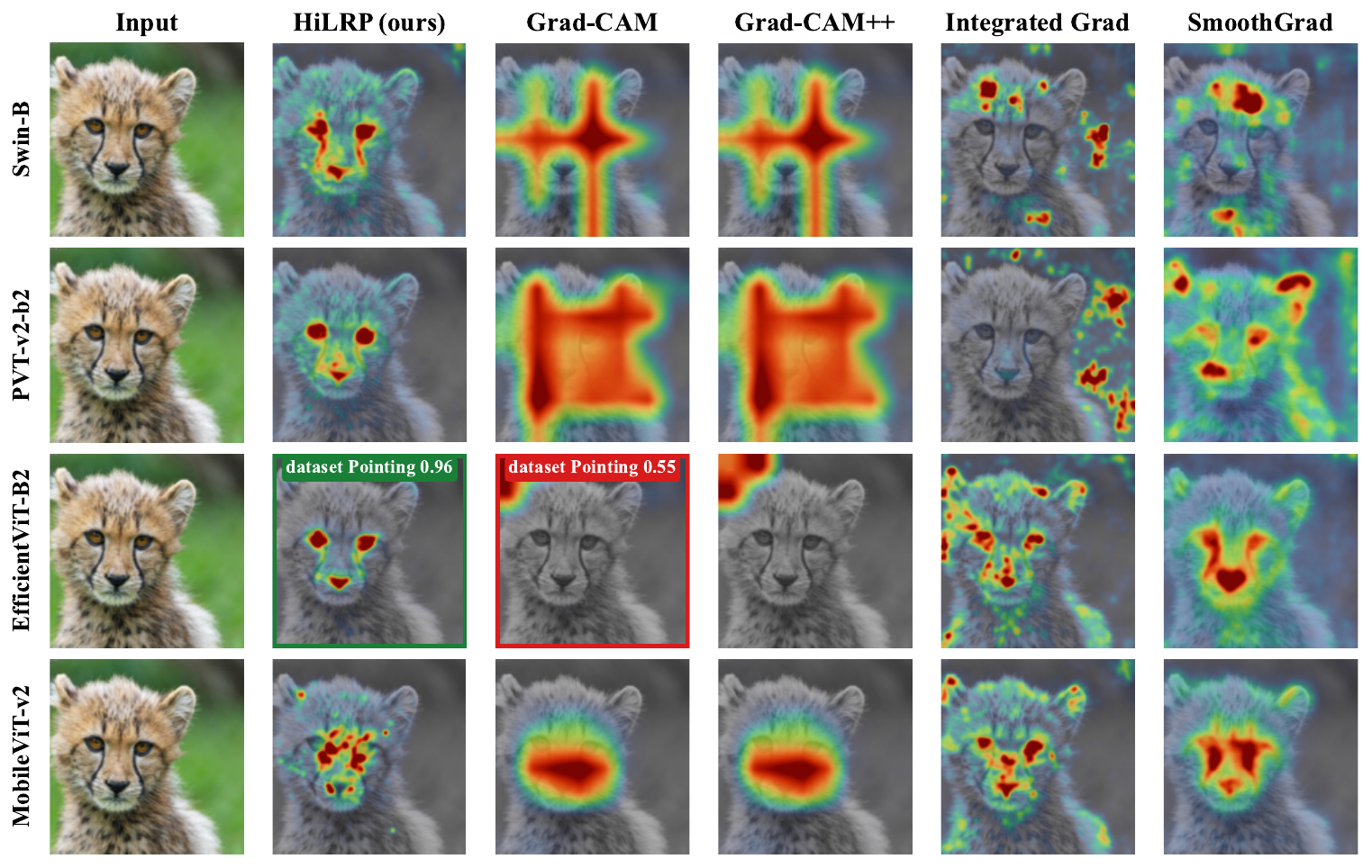}
  \caption{Attribution maps superimposed on the desaturated input, one row per architecture. Grad-CAM and Grad-CAM++ remain coherent on Swin, PVT, and MobileViT but fail to localize the object on EfficientViT-B2 (highlighted), whose linear attention provides no terminal spatial feature map. HiLRP consistently concentrates on the eyes and facial markings. Scores are dataset-level Pointing values.}
  \label{fig:arch-comparison}
\end{figure}

Attribution has not kept pace, and the resulting divergence manifests itself in two distinct ways.

The first is readily apparent. Fig.~\ref{fig:arch-comparison} applies the same methods to one image on four backbones, with architecture as the only variable. Grad-CAM~\cite{selvaraju2020gradcam} produces a coherent map on three and a near-uniform response on EfficientViT, because class activation mapping (CAM) requires a terminal spatial feature map and linear attention yields none. In the Pointing Game~\cite{zhang2018pointing}, which measures whether the highest-scoring pixel falls within the object, Grad-CAM on that backbone performs worse than a randomly placed point. Attention Rollout~\cite{abnar2020rollout} is undefined once attention is windowed, and classic layer-wise relevance propagation (LRP)~\cite{bach2015lrp} cannot be executed at all. Hyperparameter tuning does not remedy all this: each method depends on a structural property these backbones no longer possess.

The second failure is the most consequential because no symptom is visible. The principal appeal of LRP is conservation: relevance is preserved layer by layer, so the map decomposes the prediction rather than merely ranking pixels. AttnLRP~\cite{achtibat2024attnlrp} extends that guaranty through the interior of a transformer block. No rules have been published for the operations that render a backbone hierarchical, so applying it to one means falling back on raw gradients wherever a rule is absent, and the scale-invariant normalizations in those layers drive the propagated relevance toward zero. In five hierarchical backbones, we observe relevance that either collapses to exactly zero deep within the network or inflates several times (Section~\ref{sec:results}). The corresponding maps still localize well: a map summing to zero still has a well-defined maximum and still passes a localization test. According to commonly reported metrics, an invalid and a valid explanations are indistinguishable.

Together, these establish two requirements. An attribution must be \emph{defined} on the architecture being explained, and it must \emph{decompose} the prediction rather than merely rank pixels. Most methods satisfy only one. Gradient and perturbation methods apply almost anywhere but conserve nothing, so what their scores represent is unspecified. LRP and its transformer descendants~\cite{chefer2021transformer, achtibat2024attnlrp} fix that interpretation, but only on architectures for which rules have already been derived, and that cost grows with every new backbone.

We argue that this cost is avoidable. The architectures discussed above differ principally in how they \emph{arrange} operations, not in which operations they \emph{use}. Across this entire family, attention and resolution reduction decompose into just four computational primitives: linear maps (e.g., query, key, value, and output projections); bilinear products, where both factors depend on the input; normalizations and gates (e.g., softmax and LayerNorm); and reindexing operations (e.g., reshaping, window partitioning, and cyclic shifting). Windowed and spatial-reduction attention, for instance, employ these exact four, differing only in their arrangement and sparsity patterns. Because each primitive admits a single conservation-preserving relevance rule, and such operations naturally compose valid pipelines (Proposition~\ref{prop:closure}), these four rules suffice for any architecture built from them, eliminating the need for further mathematical derivation.

The hierarchical LRP (HiLRP) follows from that observation. Covering a new backbone requires only labeling each module with one of the four types; the rules and the conservation proof then follow. Patch merging, shifted-window partitioning, and spatial reduction, the three operations that motivated this work, prove to be a single problem rather than three, since each is a linear projection of a concatenated neighborhood of tokens. The guaranty is nonetheless bounded: a genuinely distinct mixing operation, such as the selective scan in state-space models, requires a fifth rule. Adding it costs one rule and leaves the others untouched, which is the principal advantage of this construction.

Evaluating these claims requires criteria that distinguish a valid attribution from an invalid one, and the common use criteria do not. Perturbation-based fidelity assumes that masking an important region alters the prediction; modern ViTs are robust enough that it often does not, and across our benchmark those scores are indistinguishable from noise. Localization presents the opposite difficulty: on datasets of large, centered objects, a fixed Gaussian prior that never examines the image saturates the Pointing Game and exceeds most genuine methods in average precision. We therefore evaluate along three axes that remain informative on these models: conservation validity, agreement with sampled Shapley values over image segments, and relevance mass within the object measured against the area a uniform map would cover (Supplementary Section~4). This paper makes four contributions.

\begin{enumerate}[leftmargin=1.2em]
  \item \textbf{Attribution from four primitives.} We reduce the attention and resolution-reduction operators of current ViTs to four operation types, each with a single conservation-preserving relevance rule, so coverage becomes a matter of construction rather than derivation. We prove conservation and conditional equivariance and verify both with respect to machine precision. Patch merging, shifted-window partitioning, and spatial reduction share one algebraic form, so a single rule and a single proof suffice for all three.

  \item \textbf{Coverage across the attention taxonomy.} We verify the decomposition on self-, cross-, and co-attention, and on windowed, linear, spatial-reduction, multi-axis, channel, dilated, and deformable attention. Eight are performed on pretrained backbones, and a cross-modal similarity score on CLIP is attributed without an additional rule. Because conservation comprises, a mechanism outside the four requires one additional rule and leaves the rest intact.

  \item \textbf{Conservation validity where prior work is degenerate.} On the windowed, spatial-reduction, multi-axis, and linear-attention families, the naive extension of attention-aware LRP either zeros or inflates relevance while continuing to score well on localization, and classic LRP cannot be executed at all. HiLRP maintains a bounded, non-degenerate decomposition on every one. Globally normalized convolutional hybrids such as MobileViT fall outside the present four and are treated as a designated extension (Section~\ref{sec:limitations}).

  \item \textbf{A cross-architecture benchmark, and an evaluation protocol that withstands it.} Across 10 architectures and 14 methods, no prior attribution is reliable on all ViT families, and Faithfulness Correlation is indistinguishable from noise in every model-method cell in which it is defined. Assessed instead by conservation validity and Shapley agreement, HiLRP is the only method that remains valid on every family, including the linear-attention backbones on which Grad-CAM fails by both criteria.
\end{enumerate}

Section~\ref{sec:related} situates the work with respect to existing attribution methods and evaluation protocols. Section~\ref{sec:methodology} develops HiLRP and its guaranties, Section~\ref{sec:setup} describes the protocol and the rationale for replacing the standard fidelity metric, and Section~\ref{sec:results} reports the results. Sections~\ref{sec:limitations} and~\ref{sec:conclusion} present the limitations and conclusions.

\section{Related work}
\label{sec:related}

The Attribution methods fall into four families, each based on architectural assumptions that a sufficiently different backbone can invalidate (Table~\ref{tab:taxonomy}). Gradient methods (Saliency~\cite{simonyan2014saliency}, Integrated Gradients~\cite{sundararajan2017ig}, Input$\times$Gradient~\cite{shrikumar2017deeplift}, SmoothGrad and VarGrad~\cite{smilkov2017smoothgrad}) differentiate the class logit with respect to the input and assume that the input gradient is smoothly behaved. CAM methods, Grad-CAM~\cite{selvaraju2020gradcam} and Grad-CAM++~\cite{chattopadhay2018gradcampp}, weight intermediate feature maps by the gradient signal and assume that a terminal spatial feature map exists to be weighted. Attention-native methods such as Attention Rollout~\cite{abnar2020rollout} read the transformer's own attention matrices, and assume that these are explicit softmax maps over a class token. Liao et al.~\cite{liao2025daam} develop this family farthest, decomposing the class token at each block to recover a per-block map and accumulating those maps to reveal how the attention of a ViT is formed from the top of the network downward. This is the closest prior work to ours, and it illustrates precisely the limitation we address: the construction is defined by the presence of a class token and explicit self-attention, so it applies only where both exist and have nothing to propagate through a patch-merging or spatial-reduction stage, and the accumulated map is not constrained to sum to the prediction. Perturbation methods (Occlusion~\cite{zeiler2014occlusion}, RISE~\cite{petsiuk2018rise}, LIME~\cite{ribeiro2016lime}, GradientSHAP~\cite{lundberg2017shap}) assume only access to the output, which makes them the most portable and the most expensive. A complementary line of work builds interpretability into the architecture rather than recovering it after training: Yu et al.~\cite{yu2023exvit} train an explainable ViT whose attention heads and attribute-guided explainer expose interpretable features directly. That approach requires retraining the backbone, whereas we explain the frozen, publicly released checkpoints practitioners deploy.

\begin{table}[!tbp]
  \centering
  \casfloatfont
  \caption{\textbf{Attribution families and their architectural assumptions.} Rows show method families and their required assumptions; columns show backbone families. A check indicates that the assumption holds, a half-disk indicates degraded or unstable behavior, and a cross indicates that the method is undefined or fails to run. HiLRP avoids these architecture-specific assumptions by operating on four conservation-preserving primitives.}
  \label{tab:taxonomy}
  \setlength{\tabcolsep}{3pt}
  \begin{tabularx}{\linewidth}{@{}llYYYYYYY@{}}
    \toprule
    \textbf{Family} & \textbf{Assumption}
    & \textbf{CNN} & \textbf{ViT} & \textbf{Swin}
    & \textbf{PVT} & \textbf{MaxV} & \textbf{MobV} & \textbf{EffV} \\
    \midrule
    Gradient
      & smooth gradients
      & \cmark & \hmark & \hmark & \hmark & \hmark & \hmark & \hmark \\
    CAM
      & spatial feature map
      & \cmark & \hmark & \cmark & \cmark & \cmark & \cmark & \xmark \\
    Attention
      & softmax $+$ \textsc{cls}
      & \xmark & \cmark & \xmark & \xmark & \xmark & \xmark & \xmark \\
    Perturbation
      & output access
      & \cmark & \cmark & \cmark & \cmark & \cmark & \cmark & \cmark \\
    Classic LRP
      & module-specific rules
      & \cmark & \xmark & \xmark & \xmark & \xmark & \xmark & \xmark \\
    \midrule
    \rowcolor{green!10}
    \textbf{HiLRP (ours)}
      & four primitives
      & \cmark & \cmark & \cmark & \cmark & \cmark & \cmark & \cmark \\
    \bottomrule
  \end{tabularx}
\end{table}

Layer-wise relevance propagation~\cite{bach2015lrp} differs in kind, in that it conserves relevance layer by layer: its output decomposes the prediction rather than ranking pixels by an unconstrained score. Montavon et~al.~\cite{montavon2017deeptaylor} placed the approach on a firmer footing through deep Taylor decomposition, deriving the propagation rules, including the $z^{+}$ rule we adopt for resolution reduction, as local Taylor expansions about a root point rather than as heuristics. Chefer et~al.~\cite{chefer2021transformer} extended this to isotropic ViTs by combining relevance with attention gradients, and AttnLRP~\cite{achtibat2024attnlrp} supplied the block-internal rules (a Taylor rule for the softmax, a uniform split for bilinear products, an identity rule for LayerNorm) that preserve conservation through a transformer block. Both conserve relevance only on flat all-to-all transformers. Neither defines a rule for the operations that render a backbone hierarchical: patch merging, shifted-window partitioning, spatial reduction, convolutional stems, or the linear-attention denominator. Applied to such a backbone, they either halt at an unsupported layer or lose conservation without warning when a scale-invariant normalization drives the propagated relevance to zero. Deriving one rule per new module does not scale with the rate at which backbones are introduced, and this gap is where HiLRP closes by construction rather than by enumeration.

Two recent studies clarify what a propagation rule must satisfy. Li et al.~\cite{li2025wblrp} show that the baseline implicit in an LRP rule is not a free choice: existing rules correspond to baselines that disregard either the model weights or the sample features, and making the baseline weight-dependent alters which evidence the resulting map credits. Their analysis is conducted on flat networks, but it reflects the same concern that motivates our treatment of the stabilizer, and it is the reason we report conservation in every setting rather than a tuned one. Vielhaben et al.~\cite{vielhaben2024vil} adopt a complementary approach, inserting virtual inspection layers so that relevance can be propagated into a representation the network never explicitly computes, such as a Fourier basis. That construction is close in spirit to our reindexing primitive: an invertible, structure-preserving transform through which relevance passes exactly, introduced without disturbing the rules on either side of it.

Evaluation presents a second major challenge. Lacking ground truth for correct attribution, the field relies on surrogate criteria. These include faithfulness metrics that perturb high-attribution regions to measure output changes~\cite{bhatt2020faithfulness}, localization metrics such as the Pointing Game~\cite{zhang2018pointing}, robustness measures~\cite{yeh2019sensitivity}, complexity measures~\cite{chalasani2020sparseness}, and the model-randomization sanity check~\cite{adebayo2018sanity}. Because these metrics frequently conflict, review studies advocate for multi-metric protocols~\cite{nauta2023anecdotal}. To ensure neutrality, we compute all metrics using \textsc{Quantus}~\cite{hedstrom2023quantus}, isolating our results from implementation bias. Although ground-truth benchmarks exist, their scope remains narrow. Datasets such as FunnyBirds~\cite{hesse2023funnybirds} and CLEVR-XAI~\cite{arras2022clevrxai} provide causal and synthetic ground truth primarily for convolutional models, along with other systematic attribution evaluations~\cite{rao2022attribution}. Although Wu et~al.~\cite{wu2024faithfulness} directly examine the faithfulness of ViT explanations, their analysis is restricted to isotropic architectures. To the best of our knowledge, no prior study has evaluated whether an attribution method remains \emph{defined} and \emph{conservation-valid} across the full spectrum of hierarchical and hybrid ViT designs under a single fixed protocol.

Two model families lie outside the primitive set treated here. State-space vision backbones mix tokens through a linear recurrence rather than a bilinear attention product, a distinct mixing primitive requiring one additional rule. Generative diffusion backbones are the second: existing explanations for them are attention- or saliency-based~\cite{park2024explaindiffusion} and rank image regions without establishing that the values returned decompose a scalar the model computes. Supplementary Section~1 provides an extended discussion of both, together with the per-paper comparisons summarized above.

\section{Methodology}
\label{sec:methodology}

\subsection{Notation and the conservation constraint}
\label{sec:notation}
Let $x$ be the input image, $f$ the frozen network, $c$ the target class, and $f_c(x)$ its logit. We denote activations as $a$ and operator outputs as $y$. \emph{Relevance} is the quantity of the core propagated backwards from the network output to the input: $R_j$ represents the portion of $f_c(x)$ attributed to activation $a_j$. The propagation initializes at the output with $R=f_c(x)$ and proceeds backward layer by layer, strictly subject to the conservation constraint at each step:

\begin{equation}
\label{eq:conservation}
\sum_{j\,\in\,\text{layer }\ell} R_j \;=\; \sum_{i\,\in\,\text{layer }\ell+1} R_i
\;=\; f_c(x),
\end{equation}
where $\ell$ indexes the depth of the network. Eq.~\eqref{eq:conservation} fundamentally distinguishes a relevance map from a saliency heuristic: it guaranties that the final attribution at the pixel level acts as a true decomposition of $f_c(x)$, preserving the prediction's units and summing precisely to its value.

HiLRP builds upon AttnLRP's block-internal attention rules~\cite{achtibat2024attnlrp} by introducing four key components: a resolution-reduction rule that unifies patch merging, shifted windows, and spatial reduction into a single coordinate-embedded linear map; a normalization guard that prevents scale-invariant norms from silently zeroing deep relevance; a conservation-preserving attention rule (CP-LRP) applied uniformly across architectural families; and a four-primitive framework that ensures coverage by construction rather than through per-architecture derivation.

\subsection{A unifying view of resolution reduction}
\label{sec:unifying}
The diverse downsampling operators employed across ViT architectures can be mathematically distilled into a single formulation. Let $\mathcal{N}=\{a^{(1)},\dots,a^{(n)}\}$ denote the ordered set of $n$ input tokens consumed during a single resolution-reduction step (e.g., $n{=}4$ for the merging of the $2{\times}2$ patch or $n{=}k^2$ for a $k{\times}k$ strided kernel), where each token $a^{(i)}\in\mathbb{R}^{d_{\mathrm{in}}}$. While patch merging concatenates and projects a $2{\times}2$ neighborhood, a strided patch-embedding convolution applies a linear map over a pixel neighborhood, and the spatial reduction in PVT and EfficientViT performs a strided convolution over the token grid. Despite these structural differences, each operation can be universally expressed as:

\begin{equation}
\label{eq:unifying}
T(\mathcal{N}) = W\,\phi\!\left(\mathrm{concat}(a^{(1)},\dots,a^{(n)})\right) = y \in \mathbb{R}^{d_{\mathrm{out}}},
\end{equation}
where $\mathrm{concat}:(a^{(1)},\dots,a^{(n)})\mapsto [a^{(1)\top},\dots,a^{(n)\top}]^\top \in \mathbb{R}^{nd_{\mathrm{in}}}$ serves as a coordinate embedding that precisely conserves relevance (Lemma~\ref{lem:conservation}), $\phi:\mathbb{R}^{nd_{\mathrm{in}}}\to\mathbb{R}^{nd_{\mathrm{in}}}$ denotes a conservation-preserving normalization function (such as the identity or LayerNorm), and $W\in\mathbb{R}^{d_{\mathrm{out}}\times nd_{\mathrm{in}}}$ represents the learned projection matrix. The output $y$ is the single merged token derived from the neighborhood $\mathcal{N}$. Consequently, Eq.~\eqref{eq:unifying} defines the fundamental operator through which HiLRP propagates relevance across any change in spatial resolution. Establishing a single propagation rule for this operator yields patch merging, strided patch embedding, and spatial reduction as natural corollaries, thereby rendering the framework fundamentally architecture-agnostic rather than a patchwork of per-model modifications. Furthermore, operations such as cyclic shifting and window partitioning are pure permutations of $\mathcal{N}$ that do not require a dedicated rule. By Lemma~\ref{lem:conservation}, permutations inherently conserve relevance, allowing the Gradient$\times$Input backend to propagate them seamlessly via native automatic differentiation.

\subsection{Attention as four conserving primitives}
\label{sec:primitives}
This principle of structural reduction extends to the attention mechanism itself, ensuring that HiLRP remains a generalized framework rather than a collection of per-architecture heuristics. Every attention mechanism analyzed in this study, ranging from global softmax to linear cross-covariance, can be decomposed into four fundamental operation classes. Because the applicable propagation rule is dictated by mathematical structure, each class is defined strictly by its algebraic form rather than its specific module implementation.

\begin{definition}[Linear map]
\label{def:linear}
An operation defined as $y_i=\sum_j w_{ij}a_j + b_i$, where the weights remain independent of the input activations. Examples include $Q$, $K$, $V$, and output projections, pooling operations, kernel feature maps, rotary and relative-position embeddings, and the resolution-reduction map formalized in Eq.~\eqref{eq:unifying}. The corresponding propagation rule is the $\epsilon$/$\gamma$-rule of Eq.~\eqref{eq:gamma-rule}, which, as shown in Lemma~\ref{lem:conservation}, conserves relevance up to the bias and stabilizer terms.
\end{definition}

\begin{definition}[Bilinear mixing]
\label{def:bilinear}
An operation $y=UV$ where \emph{both} factors are dependent on the activations, rendering the mapping non-linear with respect to either individual factor. Examples encompass the matrix products $QK^\top$ and $AV$, linear-attention products, and deformable sampling at learned offsets. The baseline propagation rule splits relevance uniformly between the two factors. However, the conservation-preserving variant employed throughout this work (Section~\ref{sec:cp}) treats one factor as a fixed gate, allowing relevance to flow exclusively through the other.
\end{definition}

\begin{definition}[Normalization and gating]
\label{def:norm}
An operation defined as $y = a \odot g(a)$ or $y = a / d(a)$, where the second factor is a scalar or per-group statistic computed directly from the activations themselves. Examples include softmax, the linear-attention denominator, LayerNorm, and squeeze-excite or sigmoid gates. The corresponding propagation rule dictates the detachment of the denominator or gate. This approach renders the operation locally linear in the numerator, which exactly conserves relevance in the detached variable.
\end{definition}

\begin{definition}[Reindexing]
\label{def:reindex}
An operation that spatially redistributes activations without combining them, meaning its transformation matrix contains exactly one nonzero entry per output. Examples consist of head splitting and merging, window partitioning, cyclic shifting, dilation, grouping, and the concatenation formalized in Eq.~\eqref{eq:unifying}. No explicit propagation rule is required for this class. According to Lemma~\ref{lem:conservation}, the ratio $z_{ij}/z_i$ is equal, ensuring that the relevance is simply copied without any leakage.
\end{definition}

\noindent Four rules suffice to cover an unbounded set of architectures because conservation is preserved under composition.

\begin{proposition}[Closure under composition]
\label{prop:closure}
Let $f = f_L \circ \cdots \circ f_1$ where each $f_\ell$ is one of
the Definitions~\ref{def:linear}--\ref{def:reindex} and each satisfies
$\sum_j R^{(\ell)}_j = \sum_i R^{(\ell+1)}_i$. Then $f$ satisfies
Eq.~\eqref{eq:conservation} end to end.
\end{proposition}

\begin{figure}[!tbp]
  \centering
  \includegraphics[width=0.76\linewidth]{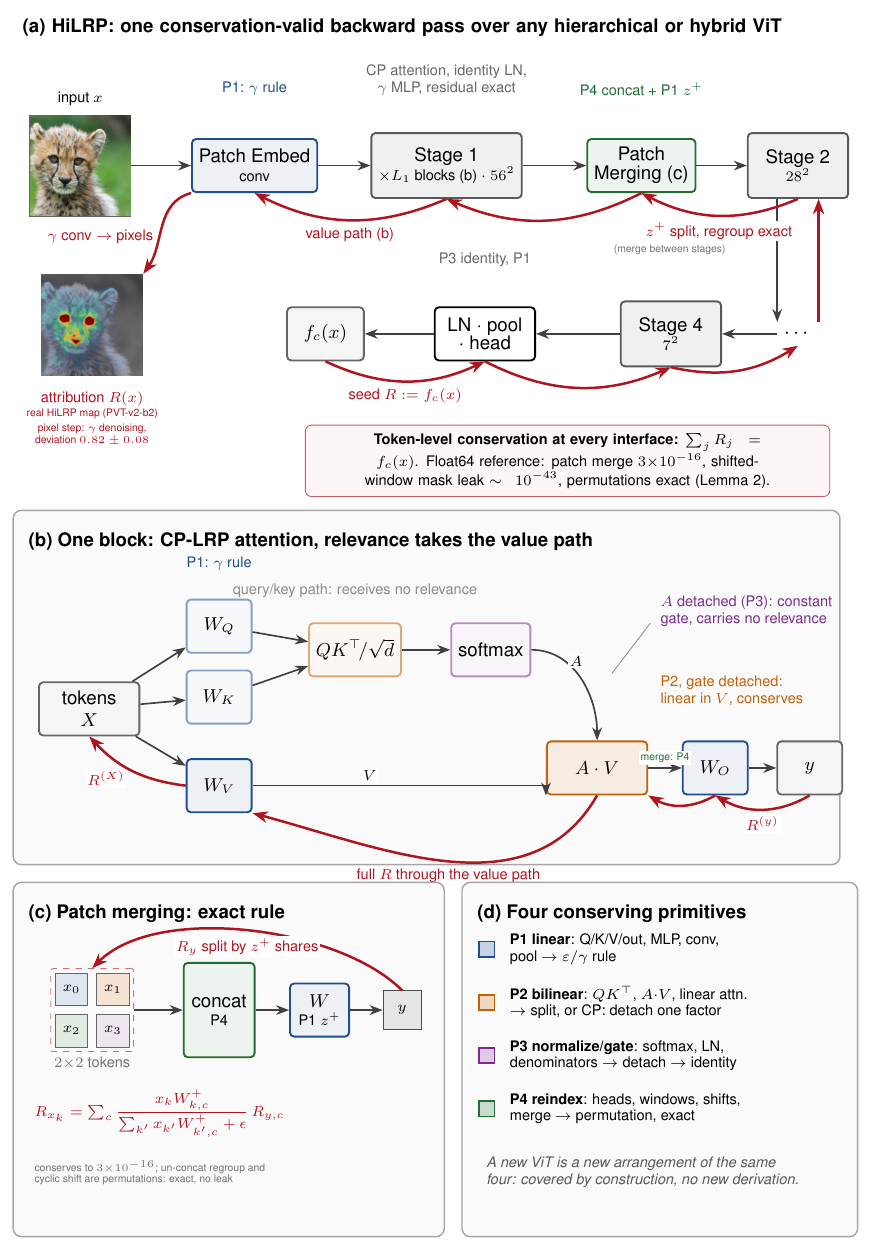}
  \caption{\textbf{HiLRP overview.} Forward propagation in black, relevance in red.
  (a) One backward pass over a hierarchical backbone.
  (b) Attention block under CP-LRP: relevance follows the value path.
  (c) Patch merging, the $z^{+}$ rule of Eq.~\eqref{eq:merge-rule}.
  (d) The four primitives and their rules.
  Colors denote primitives: linear (blue), bilinear (orange), normalization or
   gating (purple), reindexing (green).}
  \label{fig:attnlrp-propagation}
\end{figure}

\noindent The Proposition~\ref{prop:closure} converts a set of four rules into a coverage of an entire architecture family. In essence, attention consists of a linear map, followed by a bilinear mixing operation, a normalization or gate, and an aggregation over some sparsity pattern. HiLRP conserves relevance through each of these stages, and therefore through their composition. Fig.~\ref{fig:attnlrp-propagation} traces this decomposition through one multi-head attention block, coloring each operation according to the primitive to which it belongs and indicating where the relevance rule for that primitive applies.

The consequence is that coverage becomes \emph{compositional}: an attention design assembled from these primitives, in any order, inherits a conservation-valid rule with no new derivation, and a
mechanism outside them requires one added rule rather than a rederivation of the rest. Table~\ref{tab:attn-coverage} lists the span, from self-attention to linear, windowed, spatial-reduction, multi-axis, channel, and dilated attention, as well as cross- and co-attention. The two types of evidence are reported separately: seven composite types are verified in $10^{-11}$ in a float64 reference implementation, and eight are performed in pretrained models (Section~\ref{sec:results}), including cross-attention on CrossViT and cross-modal attribution in CLIP~\cite{radford2021clip}. Three types of evidence are supported by both forms.

\begin{table}[!tbp]
  \centering
  \casfloatfont
  \caption{\textbf{Attention coverage by primitive.} Primitives are linear maps (\primtag{Lin}), bilinear mixing (\primtag{Bil}), normalization or gating (\primtag{Nrm}), and reindexing (\primtag{Idx}). \primtag{F64} denotes conservation verified to $10^{-11}$ in the float64 reference suite and \primtag{Model} realization on a pretrained network; \cmark\ indicates coverage or evidence and \nd\ that the axis is not applicable. Parenthesized names are representative architectures: SE, squeeze-and-excitation; NAT, Neighborhood Attention Transformer; Def-DETR, Deformable DETR.}
  \label{tab:attn-coverage}
  \begin{tabular}{@{} l cccc c cc @{}}
    \toprule
    & \multicolumn{4}{c}{Primitives} & & \multicolumn{2}{c}{Evidence} \\
    \cmidrule(lr){2-5}\cmidrule(lr){7-8}
    Attention (example) & \primtag{Lin} & \primtag{Bil} & \primtag{Nrm} & \primtag{Idx} & & \primtag{F64} & \primtag{Model} \\
    \midrule
    Self / multi-head (ViT)    & \cmark & \cmark & \cmark & \cmark & & \cmark & \cmark \\
    Windowed / shifted (Swin)  & \cmark & \cmark & \cmark & \cmark & & \cmark & \cmark \\
    Spatial-reduction (PVT)    & \cmark & \cmark & \cmark & \cmark & &        & \cmark \\
    Linear (EfficientViT)      & \cmark & \cmark & \cmark & \cmark & & \cmark & \cmark \\
    Separable (MobileViT)$^{\dagger}$ & \cmark & \cmark & \cmark &  & &        & \cmark \\
    Multi-axis (MaxViT)        & \cmark & \cmark & \cmark & \cmark & &        & \cmark \\
    Cross-attention (CrossViT) & \cmark & \cmark & \cmark & \cmark & & \cmark & \cmark \\
    Cross-modal (CLIP)         & \cmark & \cmark & \cmark & \cmark & &        & \cmark \\
    Co-attention (bi-modal)    & \cmark & \cmark & \cmark & \cmark & & \cmark &        \\
    Channel / coordinate (SE)  & \cmark &        & \cmark &        & & \cmark &        \\
    Dilated / sparse (NAT)     & \cmark & \cmark & \cmark & \cmark & & \cmark &        \\
    Deformable (Def-DETR)      & \cmark & \cmark & \cmark & \cmark & & \cmark &        \\
    Grouped-query, RoPE        & \cmark &        &        & \cmark & & \nd    & \nd    \\
    \bottomrule
  \end{tabular}
  \\[3pt]
  \footnotesize{$^{\dagger}$MobileViT's separable-attention block decomposes into the four primitives and is realized on the pretrained model. Its surrounding single-group \textsc{GroupNorm} and channel gating fall outside them and are a designated extension (Section~\ref{sec:limitations}); the coverage claim here is for the attention block, not the whole backbone. \nd\ marks the row where both evidence axes are vacuous: grouped-query attention and rotary embeddings \emph{are} primitives, so there is nothing to verify.}
\end{table}

\subsection{Conservation-preserving attention}
\label{sec:cp}
AttnLRP propagates relevance through the softmax with a Taylor expansion of the attention matrix. This sharpens the maps obtained on isotropic ViTs but does not conserve relevance: in our ablation, it retains only $0.29$ 
of the relevance mass, so the result is a contrast-enhanced heuristic rather than a decomposition of the logit. The CP-LRP rule instead treats the attention weights as fixed gating and routes relevance through the value path alone, retaining $0.82$.

This choice acts as a simple toggle that presents a distinct trade-off: on ViT-B, the Taylor mode localizes better, but its conservation error more than doubles. Consequently, it improves visual quality, but at the cost of not properly decomposing the logit. We therefore retain CP-LRP as the conservation-preserving default across all architectural families, including flat \textsc{cls}-pooled ViTs, accepting the lower localization score that naturally results from not double-counting the \textsc{cls}-attention shortcut (Supplementary Section~2).

\subsection{Patch merging}
Hierarchical architectures such as Swin and PVT use patch merging to reduce spatial resolution while increasing the channel dimension. A patch-merging layer concatenates $2 \times 2$ neighboring patches and applies a linear projection. Let $a_{i,j}\in\mathbb{R}^{d}$ be the activation at the spatial location $(i,j)$ on the token grid before
merging, let $W\in\mathbb{R}^{d_{\mathrm{out}}\times 4d}$ be the learned projection, and let $b\in\mathbb{R}^{d_{\mathrm{out}}}$ be its bias. The merged output token at the position $(r,t)$ of the reduced grid is $y_{r,t} = W [a_{2r, 2t};\, a_{2r+1, 2t};\, a_{2r, 2t+1};\, a_{2r+1, 2t+1}] + b$, where the semicolons denote the concatenation of the four neighbors into a single vector in $\mathbb{R}^{4d}$.

Let $p\in\{1,\dots,4d\}$ index the entries of that concatenated vector, so $p$ determines both the source patch and the channel within it, and let $o\in\{1,\dots,d_{\mathrm{out}}\}$ index the output channels. Writing $R_{y,o}$ for the relevance already assigned to the output channel $o$, HiLRP applies the $z^+$-rule of the deep Taylor decomposition~\cite{montavon2017deeptaylor}:

\begin{equation}
\label{eq:merge-rule}
R_{a_p} = \sum_{o} \frac{a_p\,(W_{o,p})^+}{\sum_{p'} a_{p'}\,(W_{o,p'})^+ + \epsilon}\, R_{y,o},
\end{equation}
where $(\cdot)^{+}=\max(\cdot,0)$ denotes the positive part. Because $p'$ ranges over the same index set as $p$, the denominator sums over all contributors to the output channel $o$. By distributing relevance in proportion to these positive contributing features, the formulation prevents spatial bleeding across the merged $2 \times 2$ grid. The un-concatenation step, which routes each contribution back to its original input slot, acts as a coordinate embedding. As proven in Lemma~\ref{lem:conservation}, this embedding is exact. Consequently, the operation retains relevance entirely, except for the stabilizing $\epsilon$ terms and the bias $b$, whose constant contribution is not redistributed. During deployment, this rule is applied in its $\gamma$-stabilized form ($\gamma{=}0.25$, Section~\ref{sec:implementation}).

\subsection{Spatial reduction}
EfficientViT employs spatial reduction in its linear cross-covariance attention to reduce the cost of the 
keys and values, applying a depthwise convolution with stride $s{=}2$ over the spatial tokens. Unlike patch merging, spatial reduction is an overlapping operation. Considered per channel, it remains a linear map in a token neighborhood, so it is an instance of Eq.~\eqref{eq:unifying} and inherits 
the same rule. Let $q$ index the channels, $k$ be the size of the kernel, $a_{i,j}^q$ the scalar activation at location $(i,j)$ in channel $q$, and $W_{\mathrm{dw},q}\in\mathbb{R}^{k\times k}$ the depthwise kernel with entries indexed by the offsets $(u,v)\in\{0,\dots,k{-}1\}^2$, so that the output at position $(r,t)$ is $y_{r,t}^q=\sum_{u,v} W_{\mathrm{dw},q}^{u,v}\,a_{sr+u,\,st+v}^q$. Because the
operation overlaps, an input contributes to every output whose receptive field contains it. Let $\mathcal{O}(i,j)=\{(r,t):0\le i-sr\le k{-}1,\; 0\le j-st\le k{-}1\}$ collect those output locations; the offset linking $(i,j)$ to $(r,t)$ is $(u,v)=(i-sr,\,j-st)$. The rule is then applied
independently per channel:

\begin{equation}
\label{eq:sr-rule}
R_{a_{i,j}^q} = \!\!\sum_{(r,t)\in\mathcal{O}(i,j)}\!\!
\frac{a_{i,j}^q\,\bigl(W_{\mathrm{dw},q}^{\,i-sr,\,j-st}\bigr)^{+}}
     {\sum_{u',v'} a_{sr+u',\,st+v'}^q\,\bigl(W_{\mathrm{dw},q}^{u',v'}\bigr)^{+} + \epsilon}\;
R_{y_{r,t}^q},
\end{equation}
where $(u',v')$ range over the kernel support, so the denominator is the positive pre-activation of the output $(r,t)$ to which the numerator contributes. Summing Eq.~\eqref{eq:sr-rule} over all inputs recovers $\sum_{r,t} R_{y_{r,t}^q}$, since the numerators belonging to a given output sum up to its denominator; therefore, the step conserves relevance up to the bias and $\epsilon$ terms (Lemma~\ref{lem:conservation}). Distributing relevance in proportion to the positive contributions identifies the most active sub-regions within the overlapping windows. In deployment this rule too is applied in its $\gamma$-stabilized form ($\gamma{=}0.25$, Section~\ref{sec:implementation}).

\subsection{Theoretical guaranty}
\label{sec:guaranty}
Two properties follow from the propagation rules and can be verified independently of any dataset. Both rest on a single algebraic identity.

\begin{lemma}[Conservation by sum exchange] \label{lem:conservation}

Let $W\in\mathbb{R}^{d_{\mathrm{out}}\times d_{\mathrm{in}}}$ be a weight matrix with entries $w_{ij}$, defining the linear map
$y_i=\sum_{j=1}^{d_{\mathrm{in}}} w_{ij}a_j$, where $j\in\{1,\dots,d_{\mathrm{in}}\}$
indexes the input activations $a_j$ and $i\in\{1,\dots,d_{\mathrm{out}}\}$ indexes the outputs $y_i$. Define the \emph{contribution} of input $j$ to output $i$ as $z_{ij}=w_{ij}a_j$, and the total pre-activation at output $i$ as $z_i=\sum_{j} z_{ij}$. Then the rule
$R(a_j)=\sum_i (z_{ij}/z_i)\,R(y_i)$, which redistributes each output's relevance among its inputs in proportion to their contributions, satisfies
$\sum_j R(a_j)=\sum_i R(y_i)$. When $W$ is a $0$/$1$ permutation-like matrix so that exactly one $z_{ij}$ is nonzero per output (coordinate embeddings, reshapes), the rule is exact with no leakage.
\end{lemma}

Applied to Eq.~\eqref{eq:unifying}, the projection $W$ conserves relevance by sum exchange, the normalization $\phi$ by an identity rule, and the coordinate embedding exactly, so every resolution-reduction step conserves relevance up to the bias and stabilizer terms. Section~\ref{sec:results} verifies this numerically in a float64 reference implementation.

\begin{theorem}[Conditional equivariance]
\label{thm:equivariance}
Let $\pi:\{1,\dots,H\}\times\{1,\dots,W_{\mathrm{img}}\}\to\{1,\dots,H\}\times\{1,\dots,W_{\mathrm{img}}\}$
be a bijection on the input pixel grid, such as a cyclic translation, and write $\pi x$ for the image obtained by applying it to $x$. Let $L$ be the network depth and $F_\ell(x)$ the intermediate token representation at depth $\ell\in\{0,1,\dots,L\}$. Let $\Pi_\ell$ be the permutation that $\pi$ induces on the token index set at depth $\ell$, obtained by tracking how $\pi$ relabels spatial coordinates through the operations up to that depth. Suppose the forward pass commutes with $\pi$, that is, $F_\ell(\pi x)=\Pi_\ell F_\ell(x)$ at every depth, and the pooled logit is invariant, $f_c(\pi x)=f_c(x)$. Then HiLRP relevance is equivariant: $R(\pi x)=\pi R(x)$, where $R(x)$ is the input-level relevance map for image $x$.
\end{theorem}

\noindent This is an idealized guaranty: its premise, exact forward commutativity, holds for patch-aligned shifts on isotropic ViTs but not for deployed Swin, whose window masks are anchored to the canvas. We therefore emphasize the empirical, approximate symmetry transfer measured on real backbones (Section~\ref{sec:results}) and regard the theorem as the exact-commutativity limit of that behavior.

Equivariance is verified numerically in Section~\ref{sec:results} on an idealized cyclic model, for which the premise holds exactly. Deployed Swin does not satisfy it for nontrivial translations, since the odd window grid ($7$) does not align with the even merging grid ($2$). Where the premise holds only approximately, the guarantee degrades gradually. On pretrained Swin-T under window-multiple shifts the forward pass drifts by $7\times10^{-2}$ while HiLRP maps remain consistent at Spearman $\rho{\approx}0.90$, compared with $0.41$--$0.61$ for a gradient map, and the consistency declines as the forward drift increases. The attribution therefore tracks the model rather than the image.

\subsection{Label-free attribution}
\label{sec:ssl} Because HiLRP seeds the backward pass from a scalar and conserves relevance to it, that scalar need not be a class logit. Let $g:\mathbb{R}^{H\times W_{\mathrm{img}}\times 3}\to\mathbb{R}^{d}$ be a frozen self-supervised encoder mapping an image to its $d$-dimensional \textsc{cls} embedding, and let $x$ and $x'$ be two augmented views of the same underlying image (here a horizontal-flip pair; Supplementary Section~2). In place of $f_c(x)$ we seed the backward pass with the view-invariance similarity

\begin{equation}
\label{eq:view-sim}
s \;=\; \cos\!\big(g(x),\,g(x')\big)
  \;=\; \frac{g(x)^{\!\top} g(x')}{\lVert g(x)\rVert\;\lVert g(x')\rVert}
  \;\in[-1,1],
\end{equation}

The cosine similarity between the two embeddings, and attribute $s$ back to the pixels of $x$. This identifies \emph{the evidence supporting the representation's own invariance} without class labels. The cosine normalization requires its own identity rule: it is scale-invariant, so an in-graph norm would drive the relevance sum to zero, and we therefore detach its denominator, the term $\lVert g(x)\rVert\,\lVert g(x')\rVert$ in Eq.~\eqref{eq:view-sim}, mirroring the treatment of LayerNorm. HiLRP thereby becomes a probe of the pretraining objective itself.

\subsection{Implementation}
\label{sec:implementation}
HiLRP is realized on top of a Gradient$\times$Input relevance backend~\cite{achtibat2024attnlrp}, in which residual additions, window shifts, partitions, and the coordinate embedding of Eq.~\eqref{eq:unifying} route relevance exactly through native autograd. Explicit rules are therefore needed for only three module classes: LayerNorm and \textsc{gelu} (identity), attention (CP-LRP), and linear and convolutional layers ($\gamma$-rule).

The $\gamma$-rule is the deployed form of Eq.~\eqref{eq:merge-rule} and
Eq.~\eqref{eq:sr-rule} and is HiLRP's only tuned quantity. For a linear or convolutional layer with weights $w_{ij}$ and input activations $a_j$ it reads

\begin{equation}
\label{eq:gamma-rule}
R_j \;=\; \sum_i
\frac{a_j\big(w_{ij}+\gamma\,(w_{ij})^{+}\big)}
     {\sum_{j'} a_{j'}\big(w_{ij'}+\gamma\,(w_{ij'})^{+}\big)+\epsilon}\;R_i ,
\end{equation}
where $\gamma\ge 0$ up-weights positive weights relative to negative ones. At $\gamma{=}0$ the rule degenerates to the plain $\epsilon$-rule, which conserves relevance but is noisy at the pixel level; increasing $\gamma$ emphasizes positive evidence and denoises the map at the cost of a bias toward positive contributions. Conservation holds for every value of $\gamma$ (Supplementary Section~4), so $\gamma$ trades visual quality against sign balance rather than against validity. We adopt $\gamma{=}0.25$ globally
(Supplementary Section~4).

One implementation detail is structurally necessary rather than incidental. A scale-invariant normalization that escapes its rule patch forces the per-token relevance sum to exactly zero, corrupting every deep-stage attribution without raising an error, and \texttt{timm} defines normalization subclasses whose class-level \texttt{forward} bypasses a parent-class patch. We therefore patch by concrete class and add an audit that aborts if any normalization executes an unpatched forward pass. The single architecture-specific setting is the convolutional stabilizer $\gamma_{\text{conv}}$, which the deep convolutional stacks of the hybrid backbones require to be set below the global $\gamma$ (Supplementary Section~2).

Algorithm~\ref{alg:hilrp} presents the complete procedure. Two properties are noteworthy. It is a single forward and backward pass, costing one gradient-order evaluation whatever the depth or number of stages read out. Moreover, covering a new backbone affects only the patch map of lines 2--6, which assigns each module to one of the four primitives; nothing else is modified. This is the operational meaning of coverage by construction.

\begin{algorithm}[t]
\caption{HiLRP attribution}
\label{alg:hilrp}
\begin{algorithmic}[1]
\Require frozen network $f$; image $x$; class index $c$, or any differentiable scalar $s$; stabilizers $\gamma$, $\gamma_{\text{conv}}$, $\epsilon$ \Ensure  pixel relevance $R$; per-stage maps; conservation trace
\Statex \textbf{Assign rules by primitive} (once per architecture, class level)
\State LayerNorm, GroupNorm $\gets$ identity rule (detach the denominator)
\State \textsc{gelu} $\gets$ identity rule
\State attention $\gets$ CP-LRP rule: detach $A$ before the product $A V$
\State Linear $\gets \Gamma(\gamma)$; \; Conv2d $\gets \Gamma(\gamma_{\text{conv}})$
\State reindexing (heads, windows, shifts, merges) $\gets$ native autograd
       \Comment{exact, Lemma~\ref{lem:conservation}}
\Statex \textbf{Guard}
\State \Call{AuditNorms}{$f$} \Comment{abort if any input-statistic norm
       executes an unpatched \texttt{forward}}
\Statex \textbf{Propagate}
\State attach capture hooks at the patch embedding and each resolution stage
\State $x \gets x$ with gradient tracking enabled
\State $z \gets f(x)$
\State $s \gets z_c$ \Comment{or a label-free scalar, Eq.~\eqref{eq:view-sim}}
\State $s.\textrm{backward}()$
\Statex \textbf{Read out}
\For{each capture $(\ell,\, a_\ell)$}
  \State $R_\ell \gets a_\ell \odot \nabla_{a_\ell} s$
         \Comment{Gradient$\times$Input relevance}
  \State record $\textstyle\sum R_\ell \,/\, s$ \Comment{conservation trace}
\EndFor
\State $R \gets \sum_{\text{channels}} \big(x \odot \nabla_x s\big)$
\State \Return $R$, $\{R_\ell\}$, conservation trace
\end{algorithmic}
\end{algorithm}

\section{Experimental setup}
\label{sec:setup}
All models are frozen, publicly released \texttt{timm}~\cite{wightman2019timm} checkpoints, and no fine-tuning is applied. Inputs are processed at $224{\times}224$ using each model's own normalization statistics: supplying ImageNet statistics to a model trained on $[0,1]$ inputs severely degrades its top-1 accuracy and consequently invalidates all subsequent attributions, so per-model preprocessing is used throughout. Localization ground truth comes from the ImageNet-S~\cite{gao2022imagenets} semantic-segmentation masks, derived from ImageNet~\cite{russakovsky2015imagenet}; the principal localization comparisons are additionally validated on the PASCAL VOC~2007~\cite{everingham2010voc} test set using its object bounding boxes, giving a second, independent dataset. Baseline attributions and the perturbation-based metrics are computed with the \textsc{Quantus}~\cite{hedstrom2023quantus} toolkit so that every method and metric shares one validated implementation, while HiLRP runs on the LXT Gradient$\times$Input backend~\cite{achtibat2024attnlrp} with the resolution-reduction and CP-LRP rules of Section~\ref{sec:methodology} and $\gamma{=}0.25$ ($\gamma_{\text{conv}}{=}0.05$ on the deep convolutional stack of EfficientViT). Sample sizes are stated with each experiment: $100$ images for the full benchmark grid, $1000$ for the principal localization comparisons, $200$ for the rule ablation, $50$ ($m{=}64$ permutations) for Shapley agreement, and $30$ for the pretraining study.

Three hyperparameters are global throughout: the $\gamma$-rule stabilizer on linear and attention layers ($\gamma{=}0.25$), the lower stabilizer for the EfficientViT family's deep convolutional stack ($\gamma_{\text{conv}}{=}0.05$), and the division-by-zero stabilizer in every relevance rule ($\epsilon{=}10^{-6}$).

\paragraph{Metrics} Localization is measured by the Pointing Game~\cite{zhang2018pointing}, which determines whether the highest-scoring pixel falls within the annotated object, and by relevance mass inside the object referenced to the area a uniform map would cover. Validity is measured by conservation, the ratio of summed relevance to the explained scalar ($1.0$ is exact), and by Spearman agreement with sampled Shapley values over image segments, an axiomatic reference rather than a competing method. We also report Faithfulness Correlation, the fidelity measure most commonly used in the literature, and show in Section~\ref{sec:results} that it cannot discriminate between methods on these backbones: masking a high-attribution region barely alters the prediction on models this robust, so the score collapses into noise for every method, accurate localizers included. That is why conservation validity and Shapley agreement are our primary criteria. The full reproducibility manifest (software versions, exact \texttt{timm} tags, preprocessing, seeds) and the extended treatment of faithfulness are given in Supplementary Section~3.

\section{Results}
\label{sec:results}

\paragraph{No prior method is reliable across ViT families} Table~\ref{tab:bigbench} reports Pointing Game accuracy for 14 baseline attribution methods and HiLRP across 10 architectures, grouped into CNNs, isotropic ViTs, and hierarchical or hybrid ViTs. Every prior method has at least one family on which it fails. Grad-CAM is near-perfect on the CNNs, Swin, MobileViT, and MaxViT, but falls below the random-point prior on EfficientViT, whose linear attention leaves no terminal spatial feature map to be read. Attention Rollout and attention-gradient methods are undefined outside the isotropic column, and classic LRP cannot be executed on any modern backbone. The gradient and perturbation families apply everywhere but vary by as much as $0.3$ across architectures, with no structural basis for preferring one reading over another. HiLRP is the only method defined on every ViT column, attaining the highest cross-architecture mean and the smallest spread; its lowest score is on MobileViT-v2, the one backbone that falls outside the four primitives (Section~\ref{sec:limitations}). Localization alone is a weak discriminator here: ImageNet-S objects are large and centered, so the metric saturates for any method whose feature maps peak near the center. Supplementary Section~4 gives the per-architecture discussion, the Grad-CAM saturation analysis, and the mass-based localization measures evaluated against a model-free center-Gaussian control, which attains the highest average precision and IoU outright and thereby bounds what those two measures can convey.

\begin{table}[!tbp]
  \centering
  \casfloatfont
    \caption{\textbf{Benchmark results.} Pointing Game accuracy for HiLRP and
    14 baseline attribution methods across 10 architectures, grouped into CNNs,
    isotropic ViTs, and hierarchical/hybrid ViTs. \nd\ = not applicable or not run.
    R50 $=$ ResNet-50, CNX $=$ ConvNeXt-B, ViT $=$ ViT-B, DeiT $=$ DeiT-B,
    MoV $=$ MobileViT-v2, EV1/EV2 $=$ EfficientViT-B1/B2, MxV $=$ MaxViT-S.}
  \label{tab:bigbench}
  \setlength{\tabcolsep}{3pt}
  \begin{tabular*}{\linewidth}{@{\extracolsep{\fill}}l cc cc cccccc@{}}
    \toprule
    & \multicolumn{2}{c}{CNN} & \multicolumn{2}{c}{Isotropic ViT} & \multicolumn{6}{c}{Hierarchical / hybrid ViT} \\
    \cmidrule(lr){2-3}\cmidrule(lr){4-5}\cmidrule(lr){6-11}
    Method & R50 & CNX & ViT & DeiT & Swin & PVT & MoV & EV1 & EV2 & MxV \\
    \midrule
    Grad-CAM          & 0.98 & 1.00 & 0.93 & 0.90 & \textbf{1.00} & 0.84 & \textbf{1.00} & 0.70 & \underline{0.55} & \textbf{1.00} \\
    Grad-CAM++        & 0.97 & 0.95 & \underline{0.49} & 0.62 & \textbf{1.00} & 0.88 & \textbf{1.00} & \underline{0.44} & 0.56 & 0.80 \\
    Attn.\ Rollout   & \nd & \nd & 0.64 & \underline{0.48} & \nd & \nd & \nd & \nd & \nd & \nd \\
    Attention-grad    & \nd & \nd & 0.74 & 0.86 & \nd & \nd & \nd & \nd & \nd & \nd \\
    Saliency          & 0.82 & 0.73 & 0.65 & 0.54 & 0.62 & 0.62 & 0.90 & 0.86 & 0.66 & 0.58 \\
    Input$\times$Grad & 0.86 & 0.77 & 0.64 & 0.56 & 0.67 & 0.66 & 0.88 & 0.74 & 0.64 & 0.56 \\
    Integr.\ Grad    & 0.92 & 0.76 & 0.64 & 0.58 & 0.66 & 0.64 & 0.90 & 0.90 & 0.90 & 0.62 \\
    SmoothGrad        & 0.97 & 0.89 & 0.85 & 0.90 & 0.79 & 0.92 & 0.92 & 0.92 & 0.92 & 0.70 \\
    VarGrad           & 0.98 & 0.81 & 0.76 & 0.83 & 0.71 & 0.88 & 0.90 & 0.90 & 0.92 & 0.68 \\
    GradientSHAP      & 0.92 & 0.79 & 0.66 & 0.62 & 0.71 & 0.68 & 0.83 & 0.64 & 0.82 & 0.60 \\
    Occlusion         & 0.95 & 0.92 & 0.76 & 0.70 & 0.78 & 0.78 & 0.92 & \textbf{0.96} & 0.88 & 0.66 \\
    RISE              & 0.82 & 0.69 & 0.73 & 0.71 & 0.68 & 0.68 & 0.90 & 0.92 & 0.78 & 0.70 \\
    LIME              & 0.94 & 0.84 & 0.87 & 0.89 & 0.86 & 0.86 & 0.90 & 0.80 & 0.90 & 0.80 \\
    AttnLRP           & \nd & \nd & 0.78 & 0.84 & 0.98 & 0.80 & 0.95 & 0.84 & 0.93 & 0.89 \\
    \midrule
    \textbf{HiLRP (ours)} & \nd & \nd & 0.70 & 0.90 & 0.96 & \textbf{0.98} & 0.86 & \textbf{0.96} & \textbf{0.97} & 0.96 \\
    \bottomrule
  \end{tabular*}
\end{table}

\begin{figure}[!tbp]
  \centering
  \includegraphics[width=0.62\linewidth]{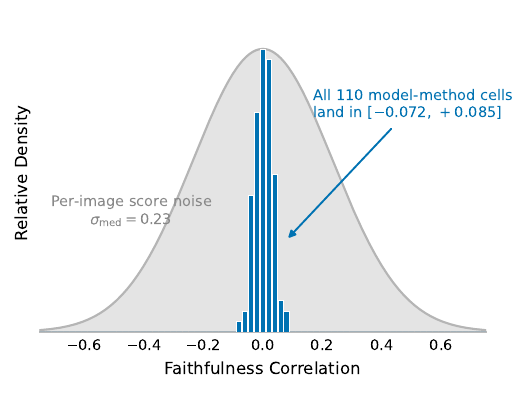}
  \caption{\textbf{Faithfulness Correlation does not discriminate between attribution methods on modern ViTs.} Per-cell mean FC over all $118$ model-method cells (blue, peak-normalized) plotted against the per-image score noise, a normal density at the median per-image $\sigma=0.23$ (gray). The entire between-method signal spans $[-0.072,+0.085]$ and lies well within the noise, so the metric does not distinguish accurate localizers from inaccurate ones.}
  \label{fig:fc-noise}
\end{figure}

\paragraph{The standard fidelity metric cannot separate the methods} Faithfulness Correlation is the measure most frequently reported in support of an attribution, yet on these backbones it is uninformative. Across every model-method cell in which it is defined, its mean lies within noise of zero and its magnitude never approaches the per-image spread (Fig.~\ref{fig:fc-noise}). The cause is architectural rather than statistical: removing a high-attribution region from a model robust to spatial masking barely alters the logit, so the correlation has nothing to register, and an accurate localizer scores no better than a poor one. Any comparison resting on this metric therefore cannot distinguish a valid explanation from an invalid one, which is why we adopt conservation validity and Shapley agreement in its place.

\paragraph{HiLRP is conservation-valid where the naive extension is degenerate} Table~\ref{tab:conservation} reports the sum of relevance in input alongside the Pointing score, so that validity and localization can be assessed together. Extending attention-aware LRP to hierarchical backbones without rules for their resolution-reduction operations produces relevance that collapses to exactly zero at depth on four of the five backbones and inflates more than threefold on the fifth, while classic LRP cannot be executed at all. The same maps still appear to be well localized. This is the failure mode of principal concern: under the metrics ordinarily reported, a degenerate attribution and a valid one are indistinguishable, because a map summing to zero still has a well-defined maximum. HiLRP maintains a bounded, non-degenerate decomposition on all five, and its token-level conservation is exact to machine precision through the proven operations. Supplementary Section~4 reports the per-stage conservation trace and the parameter-randomization sanity check, which confirms that HiLRP's maps depend on the learned weights rather than on the architecture alone.

\begin{table}[!tbp]
  \centering
  \casfloatfont
  \caption{\textbf{Conservation validity.} Pointing score / input-level relevance sum (normalized by the logit, where $1.0$ denotes exactness), $n{=}100$. The naive extension is degenerate on every backbone, being either zeroed or inflated by a factor of $3.08$, whereas HiLRP maintains a bounded, non-degenerate sum on all five attention-based backbones. The convolutional hybrid MobileViT-v2 is treated separately as an extension (Section~\ref{sec:limitations}).}
  \label{tab:conservation}
  \setlength{\tabcolsep}{14pt}
  \begin{tabular}{l cc}
    \toprule
    Backbone & Naive AttnLRP (pt / cons) & \textbf{HiLRP} (pt / cons) \\
    \midrule
    Swin-B          & 0.98 / $3.08$ (inflated) & 0.96 / $0.77$ \\
    PVT-v2-b2       & 0.80 / $0.00$ (zeroed)   & \textbf{0.98} / $0.38$ \\
    EfficientViT-B1 & 0.84 / $0.00$ (zeroed)   & \textbf{0.96} / $0.13$ \\
    EfficientViT-B2 & 0.93 / $0.00$ (zeroed)   & \textbf{0.97} / $0.35$ \\
    MaxViT-S        & 0.89 / $0.00$ (broken)   & \textbf{0.96} / $0.16$ \\
    \bottomrule
  \end{tabular}
\end{table}

\paragraph{HiLRP agrees with the axiomatic reference} Table~\ref{tab:shapley} reports Spearman agreement with sampled Shapley values over image segments, computed independently of localization. On the backbones where Grad-CAM applies, the two are statistically indistinguishable; the separation emerges exactly where Grad-CAM's spatial assumption fails, on EfficientViT, where its agreement collapses while HiLRP's holds. The claim is therefore not that HiLRP outperforms every baseline on every backbone, but that it is the only method whose agreement remains high across all of them. Localization and Shapley agreement, computed independently of one another, yield the same conclusion. Supplementary Section~4 gives the estimator's convergence, the class-sensitivity and counterfactual-class probes, the perturbation metrics and where they disagree, the rule ablation, and the $\gamma$ sweep.

\begin{table}[!tbp]
  \centering
  \casfloatfont
  \caption{\textbf{Shapley Agreement.} Spearman agreement with sampled Shapley values over image segments ($n{=}50$ images, $m{=}64$ permutations). Higher is better. Per-cell 95\% bootstrap intervals are approximately $\pm0.06$, so only the large EfficientViT gap between HiLRP ($0.417$) and Grad-CAM ($0.077$) is statistically separated; the Swin and PVT HiLRP-vs-Grad-CAM differences lie within noise, while both exceed the gradient baselines.}
  \label{tab:shapley}
  \setlength{\tabcolsep}{14pt}
  \begin{tabular}{l c c c}
    \toprule
    Method & Swin-T & PVT-v2 & EfficientViT-B2 \\
    \midrule
    \textbf{HiLRP (Ours)} & \textbf{0.528} & \textbf{0.531} & \textbf{0.417} \\
    Grad-CAM & 0.515 & 0.449 & 0.077 \\
    Integrated Grads & 0.192 & 0.279 & 0.281 \\
    SmoothGrad & 0.312 & 0.260 & 0.321 \\
    Saliency & 0.214 & 0.137 & 0.240 \\
    \bottomrule
  \end{tabular}
\end{table}

\paragraph{Cost and pretraining objectives} Two further results are reported in full in the Supplementary Material. On cost, HiLRP is a single forward and backward pass, needing one gradient-order evaluation however many stages are read out: cheaper than every multi-pass method in the benchmark, dearer than the single-pass ones, with the stage-localized maps coming from that same pass at no extra cost, where a per-stage Grad-CAM would need separate instrumentation per layer. On pretraining, because HiLRP is seeded from any differentiable scalar, a model can be explained in terms of its own representational objective rather than through an externally imposed probe; holding the backbone family and inputs fixed while varying pretraining across six public ViT-B checkpoints shows that the objective, rather than the architecture, governs what a model treats as explanatory. See Supplementary Sections~4 to~6.

\paragraph{Cross-attention and multi-modal attribution} Cross-attention, in which queries from one stream attend to keys and values from another, is handled least adequately by prior transformer LRP, yet it is a direct instance of the bilinear and softmax primitives on two input streams. CrossViT realizes it within a classification ViT: the class token of each token-scale branch attends to the tokens of the other branch through six cross-attention blocks. HiLRP applies the same CP-LRP rule to the branch self-attention and to the cross-attention. The patched model is forward-equivalent to the original (maximum absolute logit deviation $<10^{-6}$) and attains a Pointing score of $0.980$, matching the best backbone in Table~\ref{tab:bigbench} on a two-branch design for which attention rollout has no single path to follow.

\begin{figure}[!tbp]
  \centering
  \includegraphics[width=0.65\linewidth]{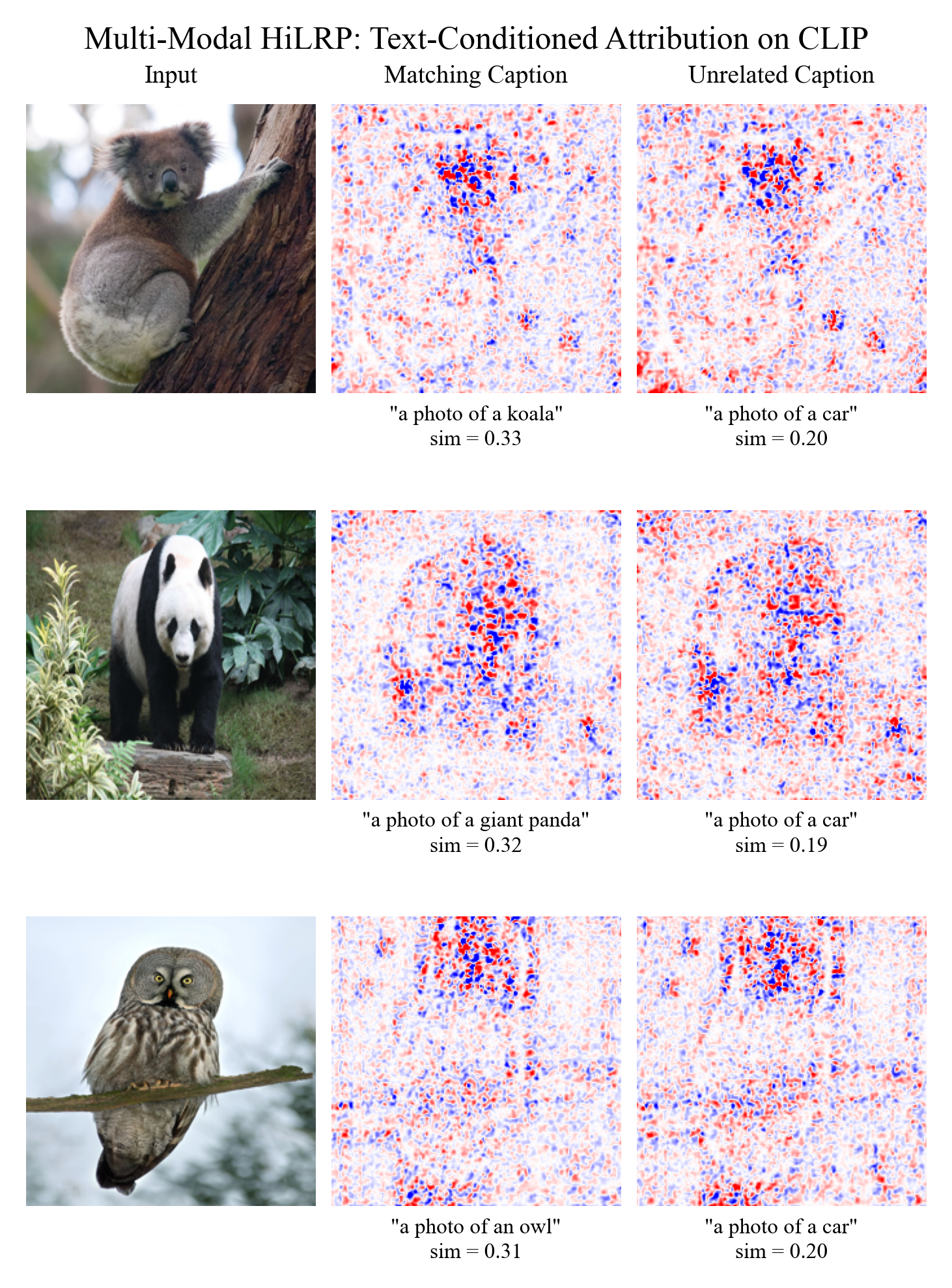}
  \caption{\textbf{Text-conditioned multi-modal attribution on CLIP.} HiLRP
  attributes the image-text similarity of Eq.~\eqref{eq:clip-sim} on CLIP
  ViT-B/16 with the caption held frozen, for a matching caption
  (\emph{``a photo of a koala/giant panda/owl''}) and an unrelated one
  (\emph{``a photo of a car''}).}
  \label{fig:clip}
\end{figure}

The scalar being explained need not originate from a single modality. On CLIP (ViT-B/16), let $f_{\text{img}}:\mathbb{R}^{H\times W_{\mathrm{img}}\times 3}\to\mathbb{R}^{d}$ be the frozen image tower, which embeds an image into the $d$-dimensional joint image-text space, and let $f_{\text{text}}:\mathcal{T}\to\mathbb{R}^{d}$ be the
frozen text tower, which embeds a tokenized caption $t\in\mathcal{T}$ into the same space. We attribute the image-text alignment
\begin{equation}
\label{eq:clip-sim}
s \;=\; \cos\!\big(f_{\text{img}}(x),\,f_{\text{text}}(t)\big),
\end{equation}
where the caption $t$ is held fixed, so no gradient flows through
$f_{\text{text}}$ and the attribution is delivered entirely to the pixels of
$x$. HiLRP thereby identifies the image evidence that supports a given textual
description. The patched image tower is forward-equivalent to the original (maximum absolute deviation $<10^{-6}$), and the attribution is text-conditioned (Fig.~\ref{fig:clip}): on clean single-object images the caption matching the image attains a higher similarity than an unrelated caption ($0.33$ against $0.20$ for the koala, $0.32$ against $0.19$ for the panda, $0.31$ against $0.20$ for the owl) and concentrates more positive relevance on the object. We present this qualitatively rather than as a localization claim, since CLIP's evidence is diffuse and its label-free localization is the weakest in Table~S4. It exercises the cross-modal scalar and the multi-head attention primitive on a pretrained multi-modal model with no new rule, extending conservation-based attribution beyond the single-model, single-label setting of prior transformer LRP.

\section{Limitations}
\label{sec:limitations}
\textbf{Conservation is exact at the token level, approximate at the pixels.} Through the operations we prove, relevance is conserved to machine precision. The $\gamma$-rule applied to the patch embedding and to convolutional stems introduces a controlled deviation, so the relevance arriving at the pixels sums to between one sixth and four fifths of the logit, depending on the backbone. We therefore state conservation as a token-level guarantee accompanied by a measured pixel-level deviation, and not as end-to-end exactness.

\textbf{MobileViT-v2 sits outside the primitive set.} Its separable-attention block decomposes into the four primitives, but the surrounding backbone does not, comprising depthwise separable convolutions, channel gating, and a single-group \textsc{GroupNorm} that normalizes over the entire spatial map. We exclude it from the conservation results, and it is the one backbone on which HiLRP ranks last. The global normalization is the mechanism: subtracting a spatial mean makes every location depend on the entire image, so a conserving attribution must carry that dependence, and the map spreads accordingly. Two of the three missing rules are evident, a per-channel rule for depthwise convolution and a detached-gate rule for channel gating; the treatment of the global normalization is not. Detaching the spatial mean, by analogy with our LayerNorm rule, degraded localization without improving conservation, and was therefore not adopted.

\textbf{Deformable attention is verified but not deployed.} It decomposes correctly in our float64 reference, since sampling at a learned offset field is detached row-stochastic mixing followed by a linear map. We have not evaluated it on a pretrained detector.

\textbf{$\gamma$ does more work than we would like.} On Swin and PVT-v2 conservation holds at every setting, so there $\gamma$ affects only the appearance of the map. On EfficientViT-B2 it does not: relevance inflates sharply in the pure-$\epsilon$ limit and recovers only at the largest setting, and our global default leaves a residual inflation. The claim that validity is setting-independent is therefore established for the windowed and spatial-reduction families, not for all of them. Localization depends on $\gamma$ everywhere, so the map a reader finally sees rests on a choice the theory does not determine. An adaptive per-layer schedule, selected under the conservation constraint, would remove this dependence.

\textbf{Equivariance holds only where the forward pass commutes.} Deployed Swin does not satisfy this condition, because its window masks are anchored to the canvas. We therefore measure the resulting approximate symmetry transfer rather than claiming exact equivariance.

\textbf{The evaluation data is easier than it looks.} ImageNet-S objects are large and centered, so the Pointing Game is permissive, and average precision and IoU are won outright by a content-free center Gaussian. Only the area-referenced energy excess discriminates, and it is computed against cached object boxes rather than pixel-accurate masks. The same single-object composition weakens the misclassified-sample probe, since a misclassification here is almost always a confusion between two classes supported by the same object. A dataset of off-center or multi-object scenes would give a cleaner signal on all of these. HiLRP also scores lower on flat isotropic models than on hierarchical ones, which is consistent with global attention diffusing relevance, although we do not isolate that mechanism experimentally.

\section{Conclusion}
\label{sec:conclusion}
Hierarchical and hybrid ViTs have rendered conservation-based attribution inapplicable. The operations that reduce resolution possessed no propagation rules, and heuristics were substituted in their place. HiLRP closes this gap by contracting the design space rather than enumerating it: the attention and resolution-reduction operators of current ViTs compose four operation types, each with a single conserving rule, so a backbone is covered by construction rather than by derivation. Patch merging, shifted-window partitioning, and spatial reduction prove to be one problem rather than three, since all are the same coordinate-embedded linear map, and a single rule accounts for all of them. We prove conservation and conditional equivariance and verify both to machine precision.

The experimental results support the same conclusion. Across the windowed, spatial-reduction, multi-axis, and linear-attention families, HiLRP is the only method that yields conservation-valid relevance where prior work is degenerate. It localizes most consistently, agrees with axiomatic Shapley values at least as closely as the strongest baseline, and is the only method that does not fail on linear attention. The same backward pass produces stage-localized maps, attributes label-free self-supervised objectives, and explains multi-modal image-text similarity, none of which requires an additional rule. We further find that Faithfulness Correlation cannot separate attribution methods on modern ViTs at all, and we argue for conservation validity and Shapley agreement in its place.

The result is bounded, and the bounds warrant explicit statement. Conservation is exact at the token level and approximate at the pixels, by a margin we measure rather than assume. One family, the globally normalized convolutional hybrids, falls outside the four primitives, and we report the resulting cost rather than tuning around it. The stabilizer $\gamma$ remains a free parameter the theory does not determine, and on linear attention it affects validity rather than appearance alone. Our evaluation data is also less demanding than the deployment settings that motivate the work: large centered objects, single-object scenes, and box-level rather than mask-level ground truth.

We anticipate that three groups will benefit. Practitioners obtain one attribution method that runs across ViT families without per-architecture code, together with a conservation check that reveals when an explanation has degenerated without warning, a failure our results show to be invisible to the metrics ordinarily reported and therefore one that a deployed pipeline would not at present detect. Architecture designers obtain a diagnostic: a block that does not decompose into the four primitives is precisely a block whose relevance semantics are undefined, and the decomposition identifies the rule that is missing. Researchers developing evaluation protocols obtain two transferable findings: that a widely used fidelity measure is uninformative on robust backbones, and that localization scores should be reported alongside the model-free control that bounds them, in the manner that detection benchmarks report a chance baseline.

Four directions follow, of which the first is the extension the framework was designed to render inexpensive. \emph{Adding primitives;} Closure under composition (Proposition~\ref{prop:closure}) renders universality an engineering objective rather than a claim: a mechanism outside the present four costs one rule and leaves every existing rule untouched, so the framework is extended rather than rederived. Three extensions are immediately tractable. Globally normalized convolutional hybrids such as MobileViT require a per-channel $\gamma$-rule for depthwise convolutions and a detached-gate rule for channel gating, together with a treatment of the global normalization that our rejected mean-detach shows is not yet settled (Section~\ref{sec:limitations}). The state-space recurrence discussed in Section~\ref{sec:related} is the second and would carry the framework beyond attention entirely. The third is the realization of our verified deformable-attention rule on a pretrained detector. Each closes a specified gap without reopening the others, which is the advantage of a compositional framework over a per-architecture one. \emph{Strengthening the evidence;} The Shapley reference should be moved from SLIC to semantic segments at larger sample counts, since the present intervals separate only the largest gaps. The whole-map localization measures of Supplementary Section~4, currently computed against cached object boxes, should be recomputed on pixel-accurate masks and repeated on off-center or multi-object scenes, for which the center-Gaussian control scores poorly by construction. \emph{Refine the rules;} An adaptive per-layer $\gamma$ schedule, selected under the conservation constraint rather than by search, would eliminate the one remaining architecture-specific setting. \emph{Widening the task.} The framework is seeded from any differentiable scalar, so dense-prediction heads for detection and segmentation, together with the denoising objective of diffusion models, are natural targets for a conservation-valid explanation.

\section*{Declaration of competing interest}
The authors declare that they have no known competing financial interests or personal relationships that could have appeared to influence the work reported in this paper.

\section*{Declaration of generative AI and AI-assisted technologies in the manuscript preparation process}
During the preparation of this work, the authors used ChatGPT to correct the sentence structure. After using this tool, the authors reviewed and edited the content as needed and take full responsibility for the content of the published article.

\section*{Data availability}
The datasets used in this study are publicly available: ImageNet-S, PASCAL VOC 2007, and the pretrained \texttt{timm} checkpoints listed in
the Supplementary Section~3. The Complete source code, configuration files, patch maps by architectures, the conservation test suite, and scripts that reproduce every table and figure will be released publicly upon acceptance.

\section*{CRediT authorship contribution statement}
\textbf{Sathiyamohan Nishankar:} Conceptualization, Methodology, Software,
Formal analysis, Investigation, Visualization, Writing -- original draft.
\textbf{Pubudu Sanjeewani:} Methodology, Supervision, Validation,
Writing -- review \& editing.
\textbf{Asanka Perera:} Supervision, Validation, Writing -- review \& editing.
\textbf{Selvarajah Thuseethan:} Supervision, Project administration,
Writing -- review \& editing.

\bibliographystyle{elsarticle-num}
\bibliography{references}

@article{nauta2023anecdotal,
  title     = {From anecdotal evidence to quantitative evaluation methods: A systematic review on evaluating explainable {AI}},
  author    = {Nauta, Meike and Trienes, Jan and Pathak, Shreyasi and Nguyen, Elisa and Peters, Michelle and Schmitt, Yasmin and Schl{\"o}tterer, J{\"o}rg and Van Keulen, Maurice and Seifert, Christin},
  journal   = {ACM Computing Surveys},
  volume    = {55},
  number    = {13s},
  pages     = {1--42},
  year      = {2023}
}

@inproceedings{rao2022attribution,
  title     = {Towards better understanding attribution methods},
  author    = {Rao, Sukrut and B{\"o}hle, Moritz and Schiele, Bernt},
  booktitle = {IEEE/CVF Conference on Computer Vision and Pattern Recognition (CVPR)},
  pages     = {10223--10232},
  year      = {2022}
}

@inproceedings{wu2024faithfulness,
  title     = {On the faithfulness of vision transformer explanations},
  author    = {Wu, Junyi and Kang, Weitai and Tang, Hao and Hong, Yuan and Yan, Yan},
  booktitle = {IEEE/CVF Conference on Computer Vision and Pattern Recognition (CVPR)},
  pages     = {10936--10945},
  year      = {2024}
}

@inproceedings{dosovitskiy2021vit,
  title     = {An Image is Worth 16x16 Words: Transformers for Image Recognition at Scale},
  author    = {Dosovitskiy, Alexey and Beyer, Lucas and Kolesnikov, Alexander and Weissenborn, Dirk and Zhai, Xiaohua and Unterthiner, Thomas and Dehghani, Mostafa and Minderer, Matthias and Heigold, Georg and Gelly, Sylvain and Uszkoreit, Jakob and Houlsby, Neil},
  booktitle = {International Conference on Learning Representations (ICLR)},
  year      = {2021}
}

@inproceedings{touvron2021deit,
  title     = {Training Data-Efficient Image Transformers \& Distillation Through Attention},
  author    = {Touvron, Hugo and Cord, Matthieu and Douze, Matthijs and Massa, Francisco and Sablayrolles, Alexandre and J{\'e}gou, Herv{\'e}},
  booktitle = {International Conference on Machine Learning (ICML)},
  year      = {2021}
}

@inproceedings{liu2022convnext,
  title     = {A {ConvNet} for the 2020s},
  author    = {Liu, Zhuang and Mao, Hanzi and Wu, Chao-Yuan and Feichtenhofer, Christoph and Darrell, Trevor and Xie, Saining},
  booktitle = {IEEE/CVF Conference on Computer Vision and Pattern Recognition (CVPR)},
  year      = {2022}
}

@inproceedings{radford2021clip,
  title     = {Learning Transferable Visual Models from Natural Language Supervision},
  author    = {Radford, Alec and Kim, Jong Wook and Hallacy, Chris and Ramesh, Aditya and Goh, Gabriel and Agarwal, Sandhini and others},
  booktitle = {International Conference on Machine Learning (ICML)},
  year      = {2021}
}

@inproceedings{bao2022beit,
  title     = {{BEiT}: {BERT} Pre-Training of Image Transformers},
  author    = {Bao, Hangbo and Dong, Li and Piao, Songhao and Wei, Furu},
  booktitle = {International Conference on Learning Representations (ICLR)},
  year      = {2022}
}

@inproceedings{liu2021swin,
  title     = {{Swin} Transformer: Hierarchical Vision Transformer Using Shifted Windows},
  author    = {Liu, Ze and Lin, Yutong and Cao, Yue and Hu, Han and Wei, Yixuan and Zhang, Zheng and Lin, Stephen and Guo, Baining},
  booktitle = {IEEE/CVF International Conference on Computer Vision (ICCV)},
  year      = {2021}
}

@inproceedings{tu2022maxvit,
  title     = {{MaxViT}: Multi-Axis Vision Transformer},
  author    = {Tu, Zhengzhong and Talebi, Hossein and Zhang, Han and Yang, Feng and Milanfar, Peyman and Bovik, Alan and Li, Yinxiao},
  booktitle = {European Conference on Computer Vision (ECCV)},
  year      = {2022}
}

@article{mehta2022mobilevitv2,
  title   = {Separable Self-Attention for Mobile Vision Transformers},
  author  = {Mehta, Sachin and Rastegari, Mohammad},
  journal = {Transactions on Machine Learning Research (TMLR)},
  year    = {2022}
}

@inproceedings{cai2023efficientvit,
  title     = {{EfficientViT}: Lightweight Multi-Scale Attention for High-Resolution Dense Prediction},
  author    = {Cai, Han and Li, Junyan and Hu, Muyan and Gan, Chuang and Han, Song},
  booktitle = {IEEE/CVF International Conference on Computer Vision (ICCV)},
  year      = {2023}
}

@misc{wightman2019timm,
  title        = {{PyTorch} Image Models},
  author       = {Wightman, Ross},
  year         = {2019},
  howpublished = {\url{https://github.com/huggingface/pytorch-image-models}}
}

@inproceedings{simonyan2014saliency,
  title     = {Deep Inside Convolutional Networks: Visualising Image Classification Models and Saliency Maps},
  author    = {Simonyan, Karen and Vedaldi, Andrea and Zisserman, Andrew},
  booktitle = {International Conference on Learning Representations (ICLR) Workshop},
  year      = {2014}
}

@inproceedings{sundararajan2017ig,
  title     = {Axiomatic Attribution for Deep Networks},
  author    = {Sundararajan, Mukund and Taly, Ankur and Yan, Qiqi},
  booktitle = {International Conference on Machine Learning (ICML)},
  year      = {2017}
}

@inproceedings{smilkov2017smoothgrad,
  title   = {{SmoothGrad}: Removing Noise by Adding Noise},
  author  = {Smilkov, Daniel and Thorat, Nikhil and Kim, Been and Vi{\'e}gas, Fernanda and Wattenberg, Martin},
  booktitle = {ICML Workshop on Visualization for Deep Learning},
  year      = {2017}
}

@inproceedings{shrikumar2017deeplift,
  title     = {Learning Important Features Through Propagating Activation Differences},
  author    = {Shrikumar, Avanti and Greenside, Peyton and Kundaje, Anshul},
  booktitle = {International Conference on Machine Learning (ICML)},
  year      = {2017}
}

@article{selvaraju2020gradcam,
  title   = {{Grad-CAM}: Visual Explanations from Deep Networks via Gradient-Based Localization},
  author  = {Selvaraju, Ramprasaath R and Cogswell, Michael and Das, Abhishek and Vedantam, Ramakrishna and Parikh, Devi and Batra, Dhruv},
  journal = {International Journal of Computer Vision (IJCV)},
  volume  = {128},
  number  = {2},
  pages   = {336--359},
  year    = {2020}
}

@inproceedings{chattopadhay2018gradcampp,
  title     = {{Grad-CAM++}: Generalized Gradient-Based Visual Explanations for Deep Convolutional Networks},
  author    = {Chattopadhay, Aditya and Sarkar, Anirban and Howlader, Prantik and Balasubramanian, Vineeth N},
  booktitle = {IEEE Winter Conference on Applications of Computer Vision (WACV)},
  year      = {2018}
}

@inproceedings{ribeiro2016lime,
  title     = {``{W}hy Should {I} Trust You?'': Explaining the Predictions of Any Classifier},
  author    = {Ribeiro, Marco Tulio and Singh, Sameer and Guestrin, Carlos},
  booktitle = {ACM SIGKDD International Conference on Knowledge Discovery and Data Mining (KDD)},
  year      = {2016}
}

@inproceedings{lundberg2017shap,
  title     = {A Unified Approach to Interpreting Model Predictions},
  author    = {Lundberg, Scott M and Lee, Su-In},
  booktitle = {Advances in Neural Information Processing Systems (NeurIPS)},
  year      = {2017}
}

@inproceedings{zeiler2014occlusion,
  title     = {Visualizing and Understanding Convolutional Networks},
  author    = {Zeiler, Matthew D and Fergus, Rob},
  booktitle = {European Conference on Computer Vision (ECCV)},
  year      = {2014}
}

@inproceedings{petsiuk2018rise,
  title     = {{RISE}: Randomized Input Sampling for Explanation of Black-box Models},
  author    = {Petsiuk, Vitali and Das, Abir and Saenko, Kate},
  booktitle = {British Machine Vision Conference (BMVC)},
  year      = {2018}
}

@inproceedings{abnar2020rollout,
  title     = {Quantifying Attention Flow in Transformers},
  author    = {Abnar, Samira and Zuidema, Willem},
  booktitle = {Annual Meeting of the Association for Computational Linguistics (ACL)},
  year      = {2020}
}

@inproceedings{chefer2021transformer,
  title     = {Transformer Interpretability Beyond Attention Visualization},
  author    = {Chefer, Hila and Gur, Shir and Wolf, Lior},
  booktitle = {IEEE/CVF Conference on Computer Vision and Pattern Recognition (CVPR)},
  year      = {2021}
}

@article{bach2015lrp,
  title   = {On Pixel-Wise Explanations for Non-Linear Classifier Decisions by Layer-Wise Relevance Propagation},
  author  = {Bach, Sebastian and Binder, Alexander and Montavon, Gr{\'e}goire and Klauschen, Frederick and M{\"u}ller, Klaus-Robert and Samek, Wojciech},
  journal = {PLoS ONE},
  volume  = {10},
  number  = {7},
  year    = {2015}
}

@article{hedstrom2023quantus,
  title   = {{Quantus}: An Explainable {AI} Toolkit for Responsible Evaluation of Neural Network Explanations and Beyond},
  author  = {Hedstr{\"o}m, Anna and Weber, Leander and Krakowczyk, Daniel and Bareeva, Dilyara and Motzkus, Franz and Samek, Wojciech and Lapuschkin, Sebastian and H{\"o}hne, Marina M-C},
  journal = {Journal of Machine Learning Research (JMLR)},
  volume  = {24},
  year    = {2023}
}

@inproceedings{bhatt2020faithfulness,
  title     = {Evaluating and Aggregating Feature-Based Model Explanations},
  author    = {Bhatt, Umang and Weller, Adrian and Moura, Jos{\'e} M F},
  booktitle = {International Joint Conference on Artificial Intelligence (IJCAI)},
  year      = {2020}
}

@article{zhang2018pointing,
  title   = {Top-Down Neural Attention by Excitation Backprop},
  author  = {Zhang, Jianming and Bargal, Sarah Adel and Lin, Zhe and Shen, Xiaohui and Brandt, Jonathan and Sclaroff, Stan},
  journal = {International Journal of Computer Vision (IJCV)},
  volume  = {126},
  number  = {10},
  pages   = {1084--1102},
  year    = {2018}
}

@inproceedings{yeh2019sensitivity,
  title     = {On the (In)fidelity and Sensitivity of Explanations},
  author    = {Yeh, Chih-Kuan and Hsieh, Cheng-Yu and Suggala, Arun Sai and Inouye, David I and Ravikumar, Pradeep},
  booktitle = {Advances in Neural Information Processing Systems (NeurIPS)},
  year      = {2019}
}

@inproceedings{chalasani2020sparseness,
  title     = {Concise Explanations of Neural Networks Using Adversarial Training},
  author    = {Chalasani, Prasad and Chen, Jiefeng and Chowdhury, Amrita Roy and Wu, Xi and Jha, Somesh},
  booktitle = {International Conference on Machine Learning (ICML)},
  year      = {2020}
}

@inproceedings{adebayo2018sanity,
  title     = {Sanity Checks for Saliency Maps},
  author    = {Adebayo, Julius and Gilmer, Justin and Muelly, Michael and Goodfellow, Ian and Hardt, Moritz and Kim, Been},
  booktitle = {Advances in Neural Information Processing Systems (NeurIPS)},
  year      = {2018}
}

@inproceedings{hesse2023funnybirds,
  title     = {{FunnyBirds}: A Synthetic Vision Dataset for a Part-Based Analysis of Explainable {AI}},
  author    = {Hesse, Robin and Schaub-Meyer, Simone and Roth, Stefan},
  booktitle = {IEEE/CVF International Conference on Computer Vision (ICCV)},
  year      = {2023}
}

@article{arras2022clevrxai,
  title   = {{CLEVR-XAI}: A Benchmark Dataset for the Ground Truth Evaluation of Neural Network Explanations},
  author  = {Arras, Leila and Osman, Ahmed and Samek, Wojciech},
  journal = {Information Fusion},
  volume  = {81},
  pages   = {14--40},
  year    = {2022}
}

@article{gao2022imagenets,
  title   = {Large-Scale Unsupervised Semantic Segmentation},
  author  = {Gao, Shanghua and Li, Zhong-Yu and Yang, Ming-Hsuan and Cheng, Ming-Ming and Han, Junwei and Torr, Philip},
  journal = {IEEE Transactions on Pattern Analysis and Machine Intelligence (TPAMI)},
  year    = {2022}
}

@article{russakovsky2015imagenet,
  title   = {{ImageNet} Large Scale Visual Recognition Challenge},
  author  = {Russakovsky, Olga and Deng, Jia and Su, Hao and Krause, Jonathan and Satheesh, Sanjeev and Ma, Sean and Huang, Zhiheng and Karpathy, Andrej and Khosla, Aditya and Bernstein, Michael and Berg, Alexander C and Fei-Fei, Li},
  journal = {International Journal of Computer Vision (IJCV)},
  volume  = {115},
  number  = {3},
  pages   = {211--252},
  year    = {2015}
}

@inproceedings{achtibat2024attnlrp,
  title     = {{AttnLRP}: Attention-Aware Layer-Wise Relevance Propagation for Transformers},
  author    = {Achtibat, Reduan and Hatefi, Sayed Mohammad Vakilzadeh and Dreyer, Maximilian and Jain, Aakriti and Wiegand, Thomas and Lapuschkin, Sebastian and Samek, Wojciech},
  booktitle = {International Conference on Machine Learning (ICML)},
  year      = {2024}
}

@article{wang2022pvtv2,
  title   = {{PVT} v2: Improved Baselines with Pyramid Vision Transformer},
  author  = {Wang, Wenhai and Xie, Enze and Li, Xiang and Fan, Deng-Ping and Song, Kaitao and Liang, Ding and Lu, Tong and Luo, Ping and Shao, Ling},
  journal = {Computational Visual Media},
  volume  = {8},
  number  = {3},
  pages   = {415--424},
  year    = {2022}
}

@article{everingham2010voc,
  title={The PASCAL Visual Object Classes ({VOC}) Challenge},
  author={Everingham, Mark and Van Gool, Luc and Williams, Christopher K I and Winn, John and Zisserman, Andrew},
  journal={International Journal of Computer Vision (IJCV)},
  volume={88},
  number={2},
  pages={303--338},
  year={2010}
}

@article{park2024explaindiffusion,
  title   = {Explaining generative diffusion models via visual analysis for interpretable decision-making process},
  author  = {Park, Ji-Hoon and Ju, Yeong-Joon and Lee, Seong-Whan},
  journal = {Expert Systems with Applications},
  volume  = {248},
  pages   = {123231},
  year    = {2024}
}

@article{liao2025daam,
  title        = {Dynamic accumulated attention map for interpreting evolution of
                  decision-making in vision transformer},
  author       = {Liao, Yi and Gao, Yongsheng and Zhang, Weichuan},
  journal      = {Pattern Recognition},
  volume       = {165},
  pages        = {111607},
  year         = {2025}
}

@article{li2025wblrp,
  title        = {{WB-LRP}: Layer-wise relevance propagation with weight-dependent
                  baseline},
  author       = {Li, Yanshan and Liang, Huajie and Zheng, Lirong},
  journal      = {Pattern Recognition},
  volume       = {158},
  pages        = {110956},
  year         = {2025}
}

@article{vielhaben2024vil,
  title        = {Explainable {AI} for time series via Virtual Inspection Layers},
  author       = {Vielhaben, Johanna and Lapuschkin, Sebastian and
                  Montavon, Gr{\'e}goire and Samek, Wojciech},
  journal      = {Pattern Recognition},
  volume       = {150},
  pages        = {110309},
  year         = {2024}
}

@article{montavon2017deeptaylor,
  title   = {Explaining Nonlinear Classification Decisions with Deep {T}aylor Decomposition},
  author  = {Montavon, Gr{\'e}goire and Lapuschkin, Sebastian and Binder, Alexander and Samek, Wojciech and M{\"u}ller, Klaus-Robert},
  journal = {Pattern Recognition},
  volume  = {65},
  pages   = {211--222},
  year    = {2017}
}

@article{yu2023exvit,
  title   = {{eX-ViT}: A Novel Explainable Vision Transformer for Weakly Supervised Semantic Segmentation},
  author  = {Yu, Lu and Xiang, Wei and Fang, Juan and Chen, Yi-Ping Phoebe and Chi, Lianhua},
  journal = {Pattern Recognition},
  volume  = {142},
  pages   = {109666},
  year    = {2023}
}

\end{document}